\documentclass[letterpaper, 10 pt, conference]{ieeeconf}  

\IEEEoverridecommandlockouts                              

\usepackage{cite}
\usepackage{amsmath,amssymb,amsfonts}
\usepackage{algorithmic}
\usepackage{graphicx}
\usepackage{textcomp}
\usepackage{xcolor}

\usepackage{graphicx} 
\usepackage{stfloats} 
\usepackage{subcaption}
\usepackage{tikz}
\usepackage{tabularx}
\usepackage{booktabs}	
\usepackage{array}
\usepackage{multirow}
\usepackage{xfrac}	

\graphicspath{{./figures/}}

\definecolor{tab_red}{RGB}{234, 107, 102}
\definecolor{tab_blue}{RGB}{126, 166, 224}
\definecolor{tab_orange}{RGB}{255, 181, 112}

\title{\LARGE \bf
	A Multi-Layered Approach for Measuring the Simulation-to-Reality Gap of Radar Perception for Autonomous Driving
}

\author{Anthony Ngo$^{1}$, Max Paul Bauer$^{1}$ and Michael Resch$^{2}$
\thanks{$^{1}$Automated Driving, Cross-Domain Computing Solutions, Robert Bosch GmbH, 70565 Stuttgart, Germany,
        {\tt\small \{Anthony.Ngo, Max.Bauer\}@de.bosch.com}}%
\thanks{$^{2}$High Performance Computing Center, University of Stuttgart, 70569 Stuttgart, Germany,
        {\tt\small resch@hlrs.de}}
\thanks{ \copyright 2021 IEEE. Accepted at the 24th IEEE International Conference on Intelligent Transportation Systems (ITSC 2021).}%
}

\begin{document}

\maketitle
\thispagestyle{empty}
\pagestyle{empty}

\begin{abstract}
	With the increasing safety validation requirements for the release of a self-driving car, alternative approaches, such as simulation-based testing, are emerging in addition to conventional real-world testing.
In order to rely on virtual tests the employed sensor models have to be validated. For this reason, it is necessary to quantify the discrepancy between simulation and reality in order to determine whether a certain fidelity is sufficient for a desired intended use. There exists no sound method to measure this simulation-to-reality gap of radar perception for autonomous driving.
We address this problem by introducing a multi-layered evaluation approach, which
consists of a combination of an explicit and an implicit sensor model evaluation. The former directly evaluates the realism of the synthetically generated sensor data, while the latter refers to an evaluation of a downstream target application.
In order to demonstrate the method, we evaluated the fidelity of three typical radar model types (ideal, data-driven, ray tracing-based) and their applicability for virtually testing radar-based multi-object tracking.
We have shown the effectiveness of the proposed approach in terms of providing an in-depth sensor model assessment that renders existing disparities visible and enables a realistic estimation of the overall model fidelity across different scenarios.
\end{abstract}


\section{Introduction}
\label{sec: introduction}
A reliable perception and understanding of the environment is essential for any autonomous vehicle. Therefore, a redundant sensor set is used to provide a high robustness and fulfill safety requirements \cite{berk_exploiting_2019}. In addition to camera and lidar, radar sensors play a key role due to their ability to directly measure the radial velocity of an object via the Doppler effect and their robustness in adverse weather conditions \cite{skolnik_radar_2008}. This allows both relevant objects to be detected and tracked over time, contributing to the general understanding of the prevailing situation \cite{schumann_scene_2020}.

The safety validation of such an autonomous system is a highly complex problem and new approaches are needed since a statistical proof of safety based on real-world testing does not scale \cite{stellet_validation_2020}.
The combination of field tests and simulation-based testing is a promising approach to substantially reduce the validation effort of autonomous driving  \cite{jesenski_simulation-based_2019}.
Several areas of application can be found in the literature for using synthetically generated sensor data in the development and testing process \cite{schlager_state---art_2020}. Sligar employs an accurate physics-based radar simulation to create a virtual sensor data set in order to train a machine learning-based object detection model \cite{sligar_machine_2020}. In contrast, Hartstern et al. use probabilistic sensor models to identify the optimal sensor setup solution in early development stages since they provide a wide range of modification parameters and adjustable settings \cite{hartstern_simulation-based_2020}. Ponn et al. utilize phenomenological sensor models to automatically create challenging and critical scenarios based on a sensor setup model of the autonomous vehicle \cite{ponn_automatic_2020}.

Given that the requirements for a sensor model can vary greatly with the intended use, the sensor model has to be validated and the right trade-off between model realism and computation time must be found. In particular, high fidelity radar simulations face the challenge of very high demands on computation time \cite{ngo_sensitivity_2020}. Although it is straightforward to measure the execution time of a simulation run, the estimation of the sensor model fidelity is quite complex since not only the sensor model itself, but also the virtual environment have to be modeled and validated \cite{rosenberger_towards_2019}.

Schlager et al. infer the fidelity of a sensor model by considering the inputs, and outputs and the modeling principle \cite{schlager_state---art_2020}. For example, a sensor model is classified as high fidelity if it uses rendering techniques such as ray tracing, which is a purely qualitative assessment and does not necessarily apply to radar simulations. Moreover, different approaches can be found in the literature that compare synthetically generated and real radar data qualitatively  \cite{owaki_hybrid_2019, deep_radar_2020, chipengo_high_2020}.
In contrast to a direct evaluation of simulated radar data, the sensor model can be evaluated indirectly by assessing the difference in outputs from a downstream algorithm fed with real and synthetic data \cite{bernsteiner_radar_2015, holder_how_2020}.
For lidar simulations, quantitative evaluations based on distance metrics and occupancy grid exist
\cite{rosenberger_benchmarking_2019, schaermann_validation_2017, browning_3d_2012}. Although lidar point clouds are comparable to radar detections, the question arises whether these techniques can be used to evaluate simulated radar data, considering that radar data is more sparse and of stochastic nature \cite{holder_measurements_2018}. 
Reway et al. propose a testing method for measuring the fidelity of synthetic video data indirectly by evaluating the object detection algorithm fed with real and simulated data \cite{reway_test_2020}.

Although many sensor simulation approaches can be found in the literature, the problem of validating and quantitatively evaluating the overall fidelity of a radar model remains unsolved.
Since it can be assumed that a deviation between real and synthetically generated data exists, it is essential to validate the radar model in order to rely on virtual tests \cite{rosenberger_towards_2019}. Novel methods are needed to measure how large the simulation-to-reality gap is with the purpose to decide whether a sensor model is suitable for an application purpose \cite{ngo_deep_2021}. 
In addition to evaluating the radar data generated, the quality of the data must also be examined with respect to its applicability for an intended use, because the requirements on the radar simulation fidelity can vary greatly depending on the application. For this reason, the performance with the desired target application should also be included in the overall assessment for a reliable and accurate measure of the overall model fidelity.

In this paper, we propose a multi-level testing method for measuring the overall simulation-to-reality gap for virtual testing of perception functions (see Fig.~\ref{fig: overview}). Our approach consists of a combination of an explicit and implicit sensor model evaluation. The former assesses the simulated sensor data directly, while the latter refers to an indirect evaluation by assessing the output of a downstream target application.
Multi-object tracking is chosen as an exemplary target application in this work, which is a typical 
use case of radar perception.
We furthermore subdivide both levels into a comprehensive (high level) and a feature level (low level) evaluation (see Fig.~\ref{fig: fidelity levels}). This division into four different fidelity levels provides a holistic sensor model assessment that makes existing disparities transparent and enables a realistic estimation of the overall model fidelity.

The remainder of the paper is structured as follows: At first, Section~\ref{sec: radar data generation and perception} introduces the generation of real and simulated data along with the radar perception. Section~\ref{sec: validation methodology} describes the proposed validation methodology with the different evaluation levels in detail. Section~\ref{sec: experiments and results} presents the conducted experiments and discusses the effectiveness of the method. Lastly, Section~\ref{sec: conclusion} concludes this paper with a brief outlook on further research.

\section{Radar Data Generation and Perception}
\label{sec: radar data generation and perception}
\begin{figure*}[t]
	\centering
	\setlength{\fboxrule}{0pt}
	\framebox{\parbox{\linewidth}{
			\includegraphics[width=\linewidth] {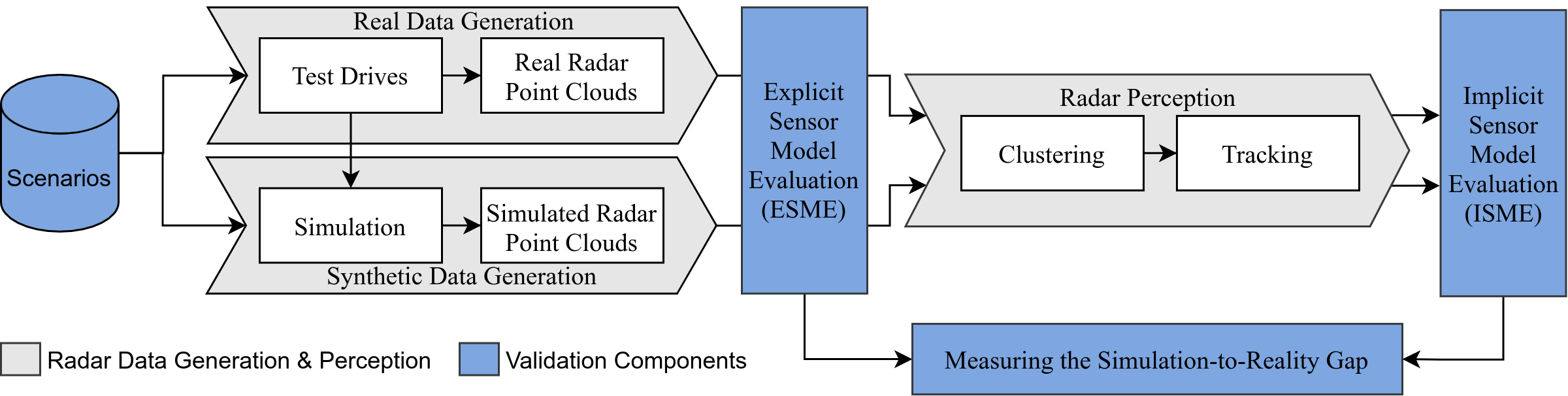}
	}}
	\caption{Overview of the proposed validation approach.}
	\label{fig: overview}
\end{figure*}

This section describes the generation of real and synthetic radar data as well as the subsequent prediction by a radar perception module. Each process consists of several steps, which are illustrated in Fig.~\ref{fig: overview} and are elaborated in the following.

\subsection{Real Data Generation}
The real data generation as a reference for comparison is an essential element for the sensor model evaluation. Since the reference data serve as the basis for the simulation, a high degree of accuracy is fundamental. In this work, the test drives are conducted on a proving ground, where the ground truth data of the objects can be obtained by using a differential global positioning system (DGPS) with an inertial measurement unit (IMU) as a reference system. The ego vehicle on which the radar sensor is mounted as well as the target vehicles are equipped with the reference system for a precise acquisition of position, orientation and velocity.

\subsection{Synthetic Data Generation and Radar Models}
The generation of synthetic radar data is divided in two steps: reproducing the real test drives in the simulation based on the recorded reference data, and the generation of a virtual scene of the environment from the radar sensor point of view, resulting in the simulated radar point cloud.
In order to demonstrate the method, three typical radar models were implemented, and they are schematically illustrated in Fig.~\ref{fig: radar models}.

\subsubsection{\textbf{Ideal Radar Model (IRM)}}
\begin{figure}[thpb]
	\centering
	\setlength{\fboxrule}{0pt}
	\framebox{\parbox{\linewidth}{
			\includegraphics[width=\linewidth] {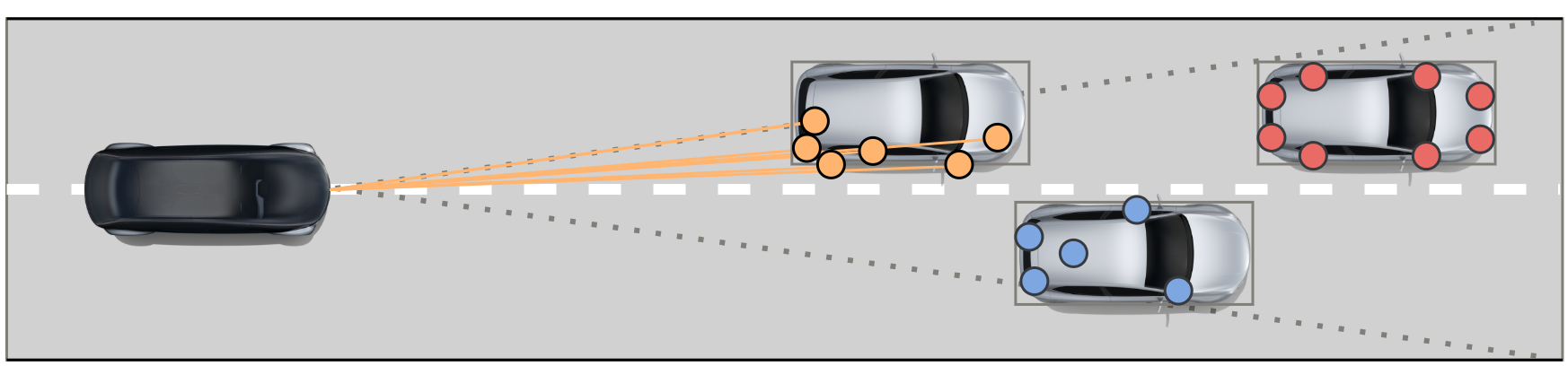}
	}}
	\caption{A schematic representation of the different radar models: ideal radar model (\protect\tikz\protect\draw[red, color=black, fill=tab_red] (0, 0) circle (2.5pt);); data-driven model (\protect\tikz\protect\draw[red, color=black, fill=tab_blue] (0, 0) circle (2.5pt);); ray tracing-based model (\protect\tikz\protect\draw[red, color=black, fill=tab_orange] (0, 0) circle (2.5pt);)}
	\label{fig: radar models}
\end{figure}

An ideal sensor model is a ground truth model, which emulate idealized behavior of the radar sensor. Such models merely consider the geometric field of view (FOV) without measurement errors or sensor specific physical effects, i.e. objects are detected any time they are within the sensor's measurement range. Due to their simplicity and fast execution, they are suitable for early testing of perception algorithms in either ideal conditions or under the assumption that sensor errors are negligible.
In order to assure a consistent sensor model output across the used sensor models in form of radar detections, the radar points of the IRM are distributed along the shell of a detected vehicle according to a simplified scattering center model \cite{schuler_extraction_2008, buddendick_bistatic_2010} (see Fig.~\ref{fig: radar models}).

\subsubsection{\textbf{Data-Driven Model (DDM)}}
Data-driven sensor models strive to approximate radar outputs by learning specific characteristics from real sensor measurements without the need to explicitly model the radar phenomena. Since these measurements inherently hold information about the observed environment, the need for rich details about the surroundings such as object material properties is eliminated.
We use as a data-driven or stochastic model a modified implementation of an existing radar sensor measurement model developed and trained from Scheel et al. \cite{scheel_tracking_2019}. Hereby, the measurement model was learned from real data using variational Gaussian mixtures. It is used to process multiple radar detections for an object in order to perform measurement-to-object associations in extended object tracking. In this work, we build on this approach in such a reversed way that we predict the radar measurements from a given object state.
We furthermore sample from the learned marginal density conditioned on the aspect angle of a detected vehicle with the purpose to generate radar detections. The number of samples is determined by a distribution conditioned on the radial distance defined by the real radar measurements.

\subsubsection{\textbf{Ray Tracing-Based Model (RTM)}}
The open-source simulator CARLA \cite{dosovitskiy_carla_2017} is used to implement the ray tracing-based radar simulation. Moreover, the testing site was virtually reproduced in the simulation and perceived by the radar model to generate simulated radar detections. The radio wave propagation is approximated using a ray tracing or rather ray casting approach based on the geometric optics diffraction theory \cite{keller_geometrical_1962}. In this context, radio waves are modeled as a bundle of rays and each beam hitting an object within the sensor's field of view returns a reflection. These reflections are further processed by calculating the signal-to-noise ratio (SNR) at each location. In general, the general performance of a radar sensor can be described by the SNR, which is defined by the received signal power and the noise power at the location of the reflection. As a final step, a detection threshold is applied in order to distinguish targets from the prevailing noise and clutter and generate radar detections. We furthermore incorporate detection probabilities to approximate the stochastic nature of radar data. The implementation details and formulas are thoroughly elaborated in \cite{ngo_sensitivity_2020}.

\subsection{Radar Perception}
The generated radar detections are further processed by a perception module, which is represented in this work by a tracking-by-detection approach as the desired target application. The module is divided into a clustering and a tracking step. 
However, since the perception algorithm is not the focus in this work and the sensor model evaluation should be generically applicable and not developed for a specific tracking approach, the tracking module is considered as a black box.

Although this work investigates the applicability of radar models for multi-object tracking, the proposed approach is not exclusively designed for this type of perception algorithm, but rather allows further abstractions to other algorithms such as classification as well. However, the metrics used in the implicit sensor model evaluation would need to be adapted to the respective use case.


\section{Validation Methodology}
\label{sec: validation methodology}
Simulation-based testing is a promising approach to reduce the testing effort of autonomous driving.
However, since it can be assumed that a certain deviation between real and simulated data exist, it is crucial to accurately quantify this gap between simulation and reality in order to enable a profound decision based on virtual tests.

\begin{figure}[thpb]
	\centering
	\setlength{\fboxrule}{0pt}
	\framebox{\parbox{0.9\linewidth}{
			\includegraphics[width=\linewidth] {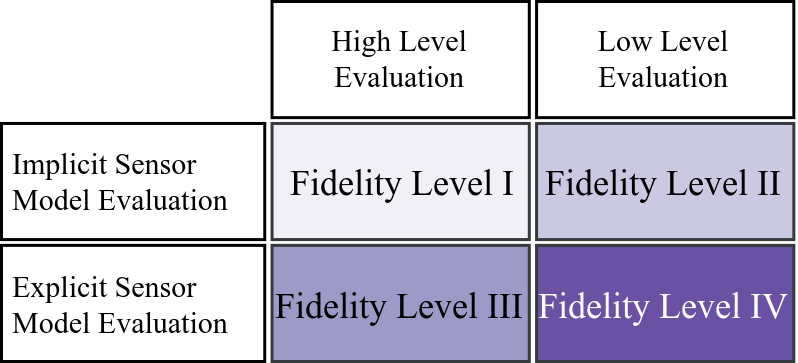}
	}}
	\caption{Fidelity Levels (FL).}
	\label{fig: fidelity levels}
\end{figure}

The methodology introduced in this section focuses on measuring this simulation-to-reality gap of a radar simulation for a specific intended use as well as rendering existing deviations visible.
We propose a multi-level testing method which consists of an explicit and an implicit sensor model evaluation. Both assessment levels are further divided into a high and low level evaluation, resulting in four different fidelity levels (FL) illustrated in Fig.~\ref{fig: fidelity levels}. 
In the following, each fidelity level is elaborated along with the quantification of the overall discrepancy between the radar simulation and the reality.

\subsection{Explicit Sensor Model Evaluation (ESME)}
The explicit or direct sensor model evaluation focuses in this work on the radar detection level, which refers to the interface after a reflection passed the detection threshold, resulting in the radar point cloud.
After the real radar data and the simulated radar data are generated, both point clouds are compared in terms of their similarity. We evaluate the characteristics of both point clouds on two different levels of detail.
The first level represents an evaluation level that uses single score metrics to assess the fidelity of the synthetically generated point clouds as a whole (high level evaluation), whereas the second level examines individual properties of the data such as number of detections independently (low level evaluation).
\subsubsection{High Level Evaluation}
Each radar detection is defined by its two-dimensional position and the Doppler velocity. 
As a first metric, we use the normalized sum of the smallest Euclidean distance from every detection in the real point cloud $X = (x_1 , ..., x_M )$ to the simulated point cloud $Y = ( y_1, ..., y_N )$, where $x_m, y_n \in \mathbb{R}^{3}$ are three-dimensional points. This \textbf{point cloud to point cloud distance} $D_{pp}$ is first introduced by Browning et al. \cite{browning_3d_2012} and is defined as follows: 
\begin{equation} \label{eq: dpp}
D_{pp}(X,Y) := \frac{1}{M} \sum_{m=1}^{M} \min_{1\leq n\leq N} ||x_m - y_n||.
\end{equation}
Moreover, the sum is divided by the respective number of points and the worst-case is assumed, since $D_{pp}$ is non-symmetrical.

The Gaussian \textbf{Wasserstein distance} $WD$ is introduced as a second metric to compare the point distributions of different radar point clouds. The Wasserstein distance measures the divergence between two distributions determined by the optimal cost of rearranging one distribution into the other. A detailed derivation of the following equation can be found in \cite{rubner_earth_2000}:
\begin{equation} \label{eq: wd}
WD(X,Y) := \frac{\sum_{m=1}^{M} \sum_{n=1}^{N} f_{m,n}d_{m,n}}{\sum_{m=1}^{M} \sum_{n=1}^{N} f_{m,n}}.
\end{equation}
Apart from the point clouds $X$ and $Y$, $m$ and $n$ represent the number of points in the point sets and the optimal cost between both distributions is represented by the optimal flow $f_{m,n}$. The Euclidean distance is chosen as the ground distance $d_{m,n}$. Consequently, $WD$ naturally extends the notion of a distance between single points to that of a distance between distributions of points.

\subsubsection{Low Level Evaluation}
To enable an in-depth evaluation and render existing deviations visible, we break down the point cloud evaluation into an assessment of specific properties individually. Each radar detection is specified in this paper by its two-dimensional location in polar coordinates (radial distance $r$, azimuth $\phi$) and the Doppler velocity. We examine each feature dimension by considering each as a probability distribution and measure the deviation across the virtual and the real domain by the \textbf{ feature specific Wasserstein distance} ($WD_{feature}$). This leads to the following metrics: $WD_r$, $WD_\phi$, $WD_{Doppler}$.
Additionally, the difference in the number of points is examined by the absolute \textbf{point number error} $PNE$.

\subsection{Implicit Sensor Model Evaluation (ISME)}
This subsection introduces the implicit or indirect evaluation of the sensor model by assessing the output of a downstream perception module fed with real and synthetic radar data.
By feeding simulated sensor data into a perception algorithm developed for real radar data and comparing both prediction results, the strengths and weaknesses of the sensor models can be assessed.
Analog to the explicit sensor model evaluation, ISME is divided into a high and a low level evaluation. The former measures the difference of both predictions overall, while the latter assesses specific properties of the perception result separately.
Since the evaluation of perception algorithms are more investigated and matured in contrast to the evaluation of radar detections, the used metrics are briefly elaborated in the following. 

\subsubsection{High Level Evaluation}
Two different metrics are used to evaluate the overall target tracking performance. The well known \textbf{optimal sub-pattern assignment} (OSPA) is used as a first metric, which was introduced by Schumacher et al. \cite{schuhmacher_consistent_2008}. OSPA consists of two components each separately accounting for localization and cardinality errors.  
Thus, OSPA metric has two adjustable parameters $p$ and $c$ that have meaningful interpretations as outlier sensitivity and cardinality penalty. In this paper, we set $p=2$ and $c=5$.
To evaluate the performance of the bounding box prediction, the \textbf{intersection over union }($IoU$) is used as the second metric. The IoU is defined as the area of the intersection between the estimated shape based on simulated radar data and the predicted shape based on real radar data, divided by the area of the union of the two shapes.

\subsubsection{Low Level Evaluation}
In order to further examine the tracking accuracy, the \textbf{root mean squared error} RMSE of the longitudinal x-position and lateral y-position estimation are calculated.
Moreover, both cardinality estimates are compared by computing the absolute \textbf{cardinality error}, while considering the estimate based on real data as ground truth.

\subsection{Measuring the Simulation-to-Reality Gap}
After assessing each level separately in the previous sections, in a last step all results are combined into one overall gap. Therefore, we propose the \textbf{simulation-to-reality gap}~$G$ that quantifies the overall discrepancy between the radar simulation and reality, which is calculated for one scenario as follows:
\begin{enumerate}
	\item select a test scenario
	\item conduct test drive and record measurements
	\item simulate test drive and generate synthetic sensor data
	\item run perception module with real and simulated sensor data
	\item perform explicit and implicit sensor model evaluation
	\item normalize metric results to the interval [0,\,1] so that zero represents the best case (no deviation)
	\item aggregate results on each fidelity level
	\item compute the average over all fidelity levels to obtain the combined simulation-to-reality gap G
\end{enumerate}

\section{Experiments \& Results}
\label{sec: experiments and results}
In this section, we examine the effectiveness of the proposed approach in terms of its ability to accurately measure the simulation-reality gap. 
First, the three radar models (IRM, DDM, RTM) introduced in Section \ref{sec: radar data generation and perception} are evaluated for one example scenario in detail with the purpose to investigate to what extent they can approximate the real radar behavior across the different fidelity levels.
Building on this, the fidelities of the sensor models are further investigated across  different scenario categories.

\subsection{Single Scenario - Qualitative Evaluation}
The radar models are first evaluated qualitatively based on their respective object tracking performance. Thereby, we investigate to which degree they can approximate the predictions based on real radar sensor data.

\begin{figure}[thpb]
	\centering
	\setlength{\fboxrule}{0pt}
	\framebox{\parbox{\linewidth}{
			\includegraphics[width=\linewidth] {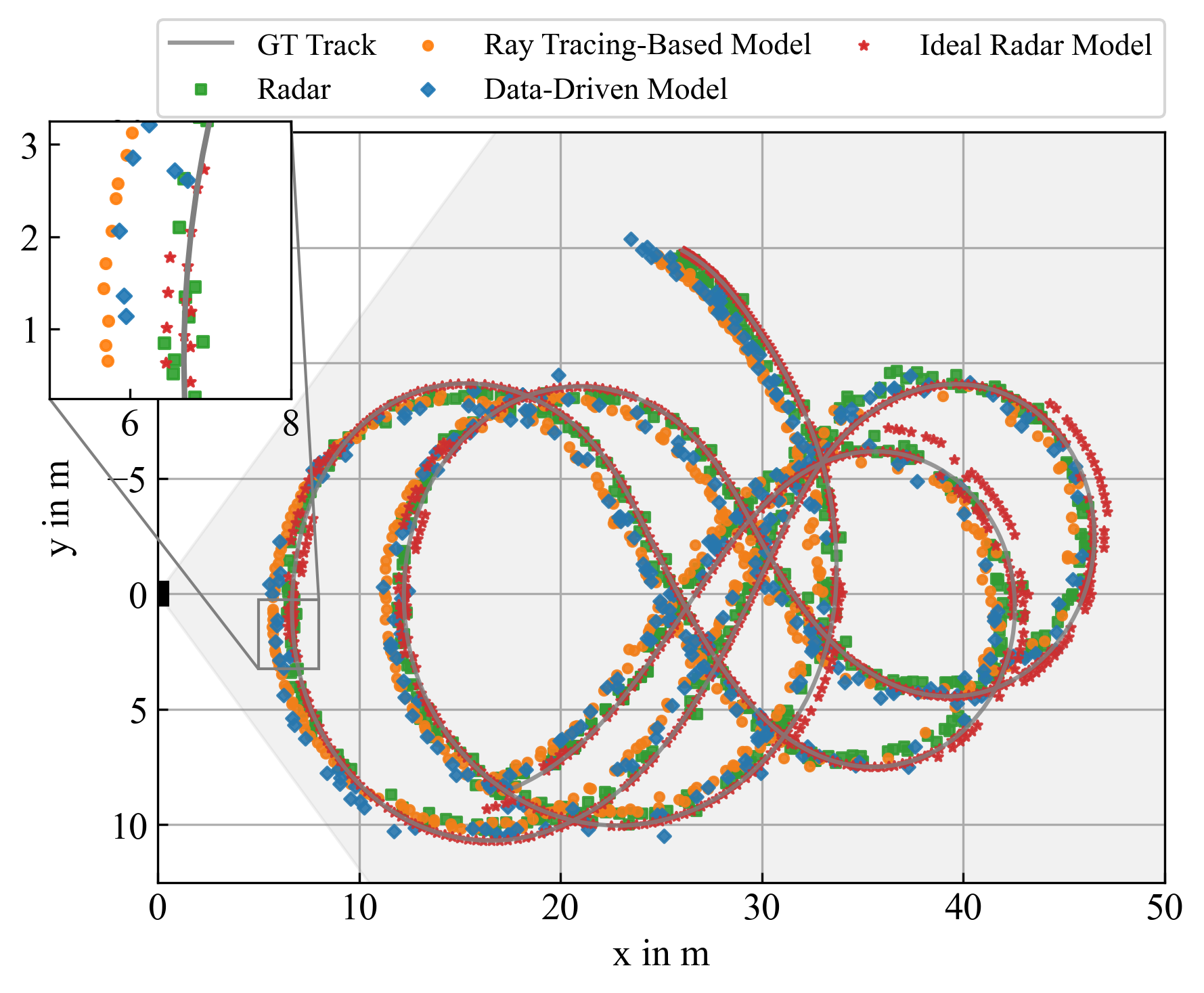}
	}}
	\caption{The color-coded points represent the radar tracking predictions with real and simulated radar sensor data, while the grey line indicates the ground truth object track (GT Track).}
	\label{fig: radar tracking}
\end{figure}

In the present example scenario, the radar sensor is static in (0,\,0) and a vehicle object drives a path in form of an eight in front of it. The results of the object tracking fed with real and synthetic radar data from each radar model are illustrated in Fig. \ref{fig: radar tracking}.

Since the tracking algorithm is parameterized with real radar data, the corresponding predicted object positions show a relatively small deviation from the ground truth track of the target object.
Due to the idealized modeling approach of the IRM, only small deviations are observed, similar to the real sensor data. These results are expected, since the radar points are distributed evenly along the bounding box of the object. Thus, the estimation of the object position is easier in comparison to the data produced by DDM and RTM, which also model the sensor error resulting in more noisy sensor data.
The latter shows a constant offset, because the simple ray casting approach is used. As a result, the radar detections are more distributed along the visible edge of the object, causing a slightly shifted predicted track.
Although similar results can be observed with the synthetically generated data by the DDM, the error observed is relatively smaller. This is due to the fact that the radar detections are distributed over the entire object rather than predominantly on the outer shell.
 
\subsection{Single Scenario - Quantitative Evaluation}
Although the IRM is an ideal sensor model, it produces in terms of object tracking prediction a more realistic estimation, i.e. similar to the real radar tracking predictions. In order to determine a realistic overall model fidelity (simulation-to-reality gap), not only the tracking prediction but also the direct output of the model, here the radar point cloud, must be considered.

We perform the implicit as well as the explicit sensor model evaluation and combine both quantitative results to a realistic estimate of the overall simulation-to-reality gap.
For a holistic sensor model assessment, multiple metrics are used at every fidelity level each assesses different characteristics in the data. The results of the metrics introduced in Section \ref{sec: validation methodology} are shown in Table \ref{tab: Evaluation results}. 
 
\begin{figure}[thpb]
	\centering
	\setlength{\fboxrule}{0pt}
	\framebox{\parbox{\linewidth}{
			\includegraphics[width=\linewidth] {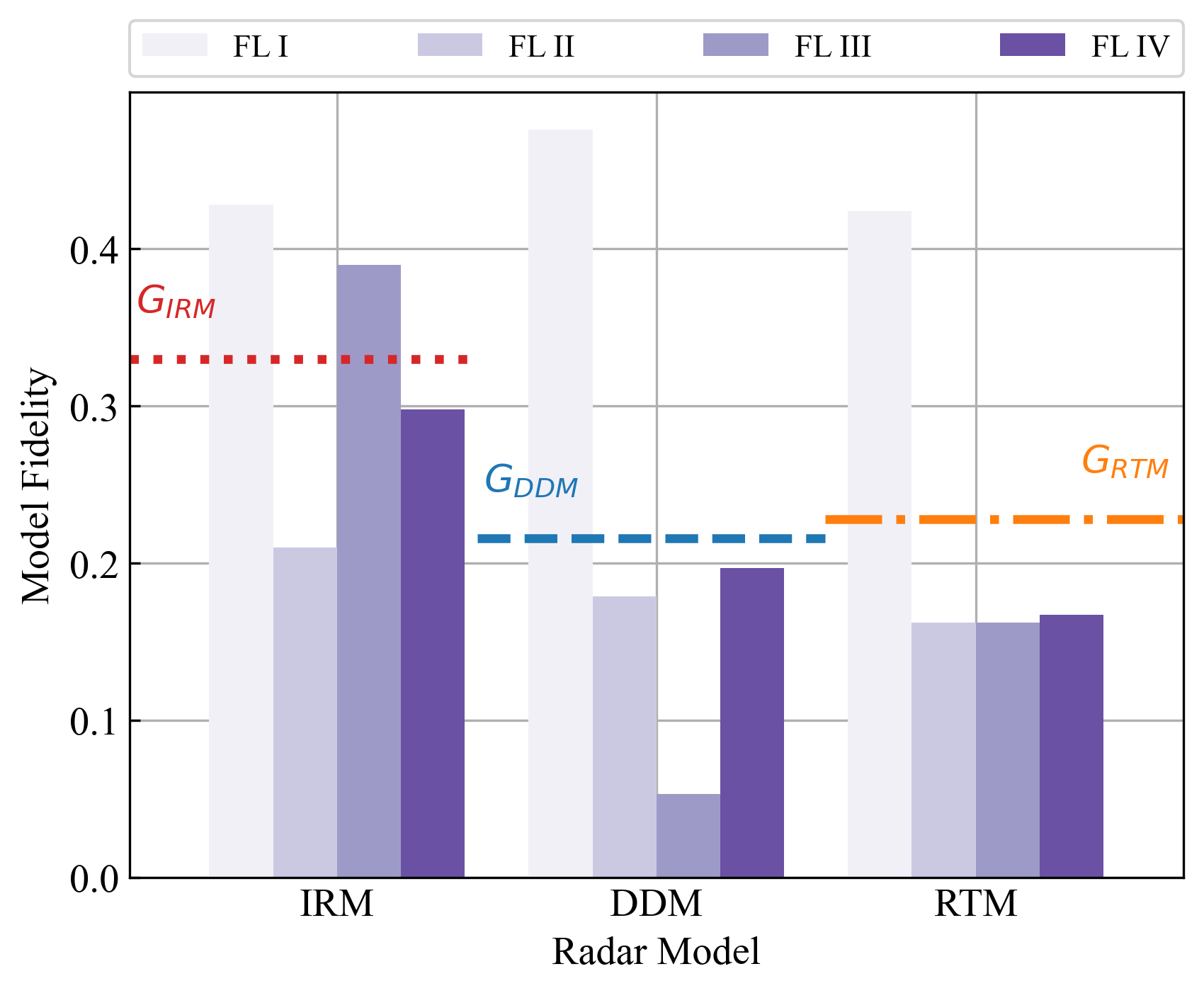}
	}}
	\caption{Aggregated results of each fidelity level along with the resulting overall simulation-to-reality gap for the `eight' scenario.}
	\label{fig: radar model fidelities}
\end{figure}

Hereby, the value of a metric represents the mean value over the scenario. Since the resulting range of values can vary widely between the different metrics, a min-max normalization is applied to rescale the range of data to [0,\,1]. By aggregating the individual results of each fidelity level, the metric results are easier to interpret, resulting in Fig. \ref{fig: radar model fidelities}. Additionally, the overall simulation-to-reality gap $G$ is illustrated allowing a quantification of the total fidelity regarding the intended use investigated.
 
\begin{table}[thpb]
	\caption{The evaluation results of the radar models across the different fidelity levels. The down arrow (resp. up arrow) indicates that the performance
		is better if the quantity is smaller (resp. greater).
	}
	\label{tab: Evaluation results}
	\begin{center}
		\begin{tabular}{l l c c c}
			\toprule
			Fidelity Level & Metric & IRM & DDM & RTM \\
			\midrule
			\multirow{2}{*}{FL I}
			& \multicolumn{1}{l}{$OSPA \downarrow$} & \multicolumn{1}{l}{0.342} & \multicolumn{1}{l}{0.314} & \multicolumn{1}{l}{0.304} \\
			& \multicolumn{1}{l}{$IoU \uparrow$} & \multicolumn{1}{l}{0.545} & \multicolumn{1}{l}{0.347} & \multicolumn{1}{l}{0.346} \\ \\
			\multirow{3}{*}{FL II}
			& \multicolumn{1}{l}{$RMSE_{x}\downarrow$} & \multicolumn{1}{l}{0.292} & \multicolumn{1}{l}{0.287} & \multicolumn{1}{l}{0.231} \\
			& \multicolumn{1}{l}{$RMSE_{y}\downarrow$} & \multicolumn{1}{l}{0.243} & \multicolumn{1}{l}{0.258} & \multicolumn{1}{l}{0.189} \\
			& \multicolumn{1}{l}{$Cardinality~Error\downarrow$} & \multicolumn{1}{l}{0.093} & \multicolumn{1}{l}{0.0} & \multicolumn{1}{l}{0.004} \\ \\
			\multirow{2}{*}{FL III}
			& \multicolumn{1}{l}{$DPP\downarrow$} & \multicolumn{1}{l}{0.381} & \multicolumn{1}{l}{0.051} & \multicolumn{1}{l}{0.152} \\
			& \multicolumn{1}{l}{$WD\downarrow$} & \multicolumn{1}{l}{0.398} & \multicolumn{1}{l}{0.055} & \multicolumn{1}{l}{0.173} \\ \\
			\multirow{4}{*}{FL IV}
			& \multicolumn{1}{l}{$PNE\downarrow$} & \multicolumn{1}{l}{0.435} & \multicolumn{1}{l}{0.304} & \multicolumn{1}{l}{0.064} \\
			& \multicolumn{1}{l}{$WD_{r}\downarrow$} & \multicolumn{1}{l}{0.26} & \multicolumn{1}{l}{0.341} & \multicolumn{1}{l}{0.331} \\
			& \multicolumn{1}{l}{$WD_{\phi}\downarrow$} & \multicolumn{1}{l}{0.089} & \multicolumn{1}{l}{0.102} & \multicolumn{1}{l}{0.112} \\
			& \multicolumn{1}{l}{$WD_{Doppler}\downarrow$} & \multicolumn{1}{l}{0.409} & \multicolumn{1}{l}{0.04} & \multicolumn{1}{l}{0.161} \\
			\bottomrule
		\end{tabular}
	\end{center}
\end{table}

It is evident that the qualitative assessment by visual comparison is reflected in the metric results, in the sense that the IRM shows in comparison to the other models a better performance in the implicit than the explicit evaluation. However, due to the relatively very large deviations in the explicit comparison, the ideal model has the highest overall gap.
In contrast, for the data-driven model and the ray tracing-based model a significant difference between the FL~I and the other levels is clearly noticeable. This is due to a large difference in the bounding box estimation, which is indicated by a poor IoU score.
Except for FL~II, the results of the remaining fidelity levels are very similar, which is why the total deviation of DDM and RTM are almost the same.

\subsection{Evaluation Across Multiple Scenarios}
In contrast to the evaluation of one specific scenario, this section extends the fidelity assessment of each model with respect to different categories of scenarios. Since the output of the sensor data as well as the performance of the object tracking can vary greatly depending on the tested situation, different categories are considered.
The scenarios can be mainly divided in single (s) and multi-object (m) scenarios (see Table~\ref{tab: scenarios}). Furthermore, typical challenging scenarios are chosen to demonstrate the proposed method. The selected scenarios do not claim to cover all relevant scenarios, but rather have an exemplary character.

\begin{table}[thpb]
	\caption{List of tested scenarios and their description.}
	\label{tab: scenarios}
	\begin{center}
		\begin{tabular}{l l}
			\toprule
			Scenario Name & Description\\
			\midrule
			oncoming\textsubscript{s}	&	target enters sensor FOV in far range								 \\
			overtake\textsubscript{s}	&	target enters FOV in near range and overtakes ego										  \\
			leading\textsubscript{s}	& ego follows a leading target															 \\
			eight\textsubscript{s}		& target drives an eight in front of static ego                    \\
			occlusion\textsubscript{m}	& multiple targets are occluded													\\
			leading\textsubscript{m} &	ego follows multiple leading targets driving in parallel   					    \\
			overtake\textsubscript{m} &	multiple targets overtake ego 													  \\
			crossing\textsubscript{m} &	multiple targets cross in front of ego 										   \\
			\bottomrule
		\end{tabular}
	\end{center}
\end{table}

\begin{figure}[thpb]
	\centering
	\setlength{\fboxrule}{0pt}
	\framebox{\parbox{\linewidth}{
			\includegraphics[width=\linewidth] {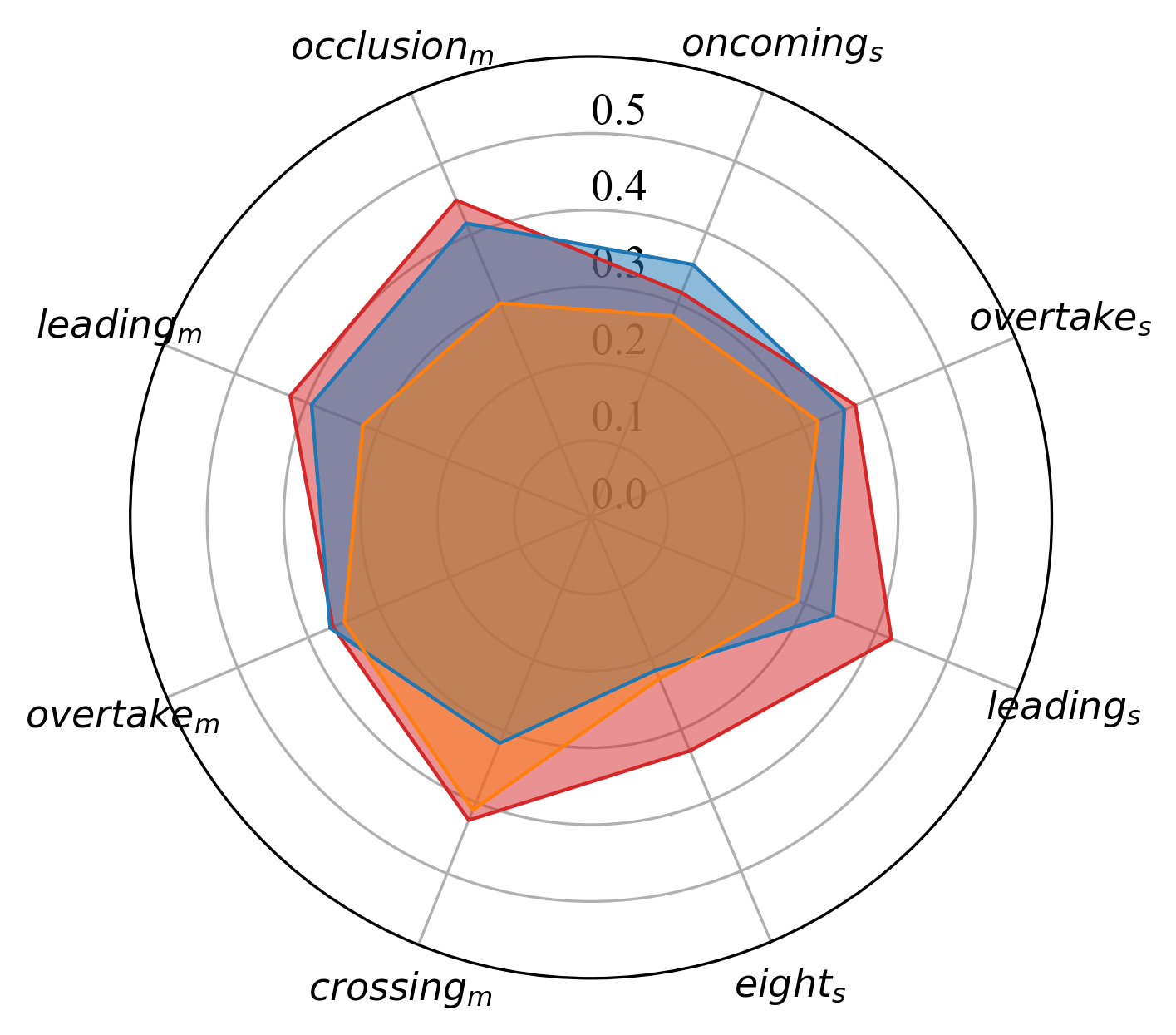}
	}}
	\caption{The simulation-to-reality gap of each radar model across multiple scenarios with 		IRM(\protect\tikz\protect\draw[red, color=tab_red, fill=tab_red, opacity=0.8] (0, 0) rectangle (0.3,0.2);), DDM(\protect\tikz\protect\draw[red, color=tab_blue, fill=tab_blue, opacity=0.8] (0, 0) rectangle (0.3,0.2);), RTM(\protect\tikz\protect\draw[red, color=tab_orange, fill=tab_orange, opacity=0.8] (0, 0) rectangle (0.3,0.2);).}
	\label{fig: gap}
\end{figure}

The results of the respective simulation-to-reality gaps are illustrated in a radar chart (see Fig.~\ref{fig: gap}). It is apparent that the ray tracing-based model achieves the lowest gap in almost all tested scenarios.
However, the crossing scenario is an exception, here both RTM and IRM perform relatively poorly, while DDM has the smallest deviation from the real data.
Furthermore, it can be seen that the deviations in the scenarios with multiple vehicles are larger than in the single-object scenarios. 
This is probably because the radar models used are relatively simplistic and therefore do not take into account effects such as multi-path reflections.

\subsection{Discussion}
Furthermore, a possible improvement exists in the metric calculation, because we might lose some information by averaging the metric results over a scenario run. It needs to be investigated to what degree a deeper analysis of the time series can improve the overall estimation result.
We assumed that the metrics on each fidelity level are weighted equally. This
is reasonable, but has to be investigated in detail in future work. 
Analog to the weighting of the metrics, we assumed that each fidelity level is of equal importance, but depending on the use case, a similar perception result might be more preferred than an accurate radar point cloud and vice versa. Therefore, it is necessary to find out which model quality requirements are needed by the different applications purposes.

\section{Conclusion}
\label{sec: conclusion}
In this paper, we introduced a multi-layered evaluation approach to measure the gap between a radar simulation and the reality. In order to investigate the effectiveness of the proposed approach, three different radar sensor models were implemented and evaluated. 
We have shown that by introducing different evaluation levels it is possible to render existing deviations in detail visible as well as assessing the sensor model fidelity across different scenarios.
This objective and quantitative evaluation enables a scaled number of tests to be performed automatically, as well as  providing the basis for an informed decision based on virtual tests, thus reducing the validation effort of autonomous driving functions.

In future work, alternative radar perception algorithms such as classification can be considered. Thereby, only the metrics of the implicit sensor model evaluation have to be adapted.
The multi-layered evaluation approach could also be applied to other sensor modalities like lidar sensors, since lidar point clouds are comparable to radar detections.

\addtolength{\textheight}{0cm}   

\bibliography{MyLibrary}
\bibliographystyle{IEEEtran}

\end{document}